\documentclass[conference]{IEEEtran}
\usepackage[utf8]{inputenc}
\usepackage{cite}
\usepackage{amsmath,amssymb,amsfonts,capt-of}
\usepackage{graphicx}
\usepackage{textcomp}
\usepackage{xcolor}
\usepackage{algorithm}
\usepackage{algpseudocode} 
\usepackage{url} 
\usepackage{float}
\usepackage{url}
\usepackage{flushend}
\usepackage{tikz}

\begin{document}

\title{Local Optimization Often is Ill-conditioned in Genetic Programming for Symbolic Regression}

\author{\IEEEauthorblockN{Gabriel Kronberger}
	\IEEEauthorblockA{
          \textit{Josef Ressel Center for Symbolic Regression} \\
          \textit{Heuristic and Evolutionary Algorithms Lab} \\
		\textit{University of Applied Sciences Upper Austria, Hagenberg Campus} \\
		Gabriel.Kronberger@fh-ooe.at ORCID: 0000-0002-3012-3189}
}

\maketitle

\begin{abstract}
  Gradient-based local optimization has been shown to improve results
  of genetic programming (GP) for symbolic regression. Several
  state-of-the-art GP implementations use iterative nonlinear least
  squares (NLS) algorithms such as the Levenberg-Marquardt algorithm
  for local optimization. The effectiveness of NLS algorithms depends
  on appropriate scaling and conditioning of the optimization
  problem. This has so far been ignored in symbolic regression and GP
  literature. In this study we use a singular value decomposition of
  NLS Jacobian matrices to determine
  the numeric rank and the condition number. We perform experiments
  with a GP implementation and six different benchmark datasets. Our
  results show that rank-deficient and ill-conditioned Jacobian
  matrices occur frequently and for all datasets. The issue is less
  extreme when restricting GP tree size and when using many non-linear functions in the function set.
\end{abstract}


\section{Introduction}
Genetic programming (GP) is an evolutionary algorithm for generating computer programs that solve given problems \cite{Koza1992,Langdon2002}. GP has been shown to be particularly effective for symbolic regression (SR). The aim of SR is to find an expression that fits a given dataset. A common objective function to be minimized for regression is the sum of squared errors of the prediction function $f(X,\theta)$
\begin{equation}
  \| y - f(X,\theta) \|_2^2,
  \label{eqn:ssr}
\end{equation}
whereby  $y$ is the target vector of length $n$, $X$ is the $n\times d$ matrix of input values, and $\theta$ is the parameter vector. Most regression methods assume a fixed structure of the prediction function and optimize the parameter vector to minimize Equation~\ref{eqn:ssr}. In contrast, SR aims to find an expression together with its parameters. For this process, the SR algorithm assembles an expression by using elements from a library of operators, functions, and terminals. When using GP for SR, this is accomplished by a stochastic evolutionary process.

Several contemporary GP implementations for SR use a memetic approach \cite{neri2011handbook} and combine evolutionary search for the structure of the model with local optimization of the real-valued parameters \cite{Kommenda2020,Burlacu2020,dick2020,virgolin:2022:SymReg,aldeia:2022:SymReg,LaCava:GPEM}. The advantage of separating the optimization problem into two sub-problems is that we can use well-established nonlinear least squares (NLS) algorithms (such as the Levenberg-Marquardt algorithm \cite{Levenberg,Marquardt}) to locally optimize the numeric parameters more efficiently than would be possible with an evolutionary algorithm \cite{topchy2001faster,trujillo2014evaluating,zflores2014, trujillo2018local,Kommenda2020}. This takes the burden of finding optimal parameters from GP which is still required to evolve the expression structure. It has been shown that extracting the parameters from the expression, performing a few iterations of NLS, and writing back the parameters can improve GP results \cite{Kommenda2020}. 

In the following we briefly summarize the principle of local optimization algorithms for NLS using the nomenclature from \cite{minpackuserguide}.
The NLS optimization problem is
\begin{equation}
 \arg\min_x  \| F(x) \|_2  .
\end{equation}
In the case of nonlinear regression for target $y$ and inputs $X$ we have $F(x) = y - f(X,x)$.
Local NLS algorithms approach this problem by starting with an initial point $x_0$ and making steps $p$ to find new points $x_{i+1}=x_{i} + p$ with $\|F(x_{i+1})\|_2 < \|F(x_{i})\|_2$. For this, a linear approximation can be used based on the assumption that $F(x + p) \approx F(x) + J(x)p$, where $J(x)$ is the Jacobian of $F(x)$. In principle this means that for each step we have to solve the linear least squares problem
\begin{equation}
  \min_{p} \|J(x)p + F(x) \|_2\, .
\end{equation}

Trust-region algorithms such as the Levenberg-Marquardt algorithm improve the convergence rate by constraining the step-size 
\begin{equation}
  \min_{p} \|J(x)p + F(x) \|_2 \quad \text{s.t.}\, \|Dp\|_2 < \Delta
  \label{eq:LevMar}
\end{equation}
whereby the maximum step size $\Delta$ is automatically adapted by the algorithm and $D$ is a diagonal scaling matrix.

Equation \eqref{eq:LevMar} is solved repeatedly to find the steps $p$ until the algorithm converges to a local optimum.
The linear least squares problem (LLS) is ill-conditioned when $J(x)$ is ill-conditioned. Ill-conditioning occurs when $J(x)$ is rank-deficient, meaning that some columns of $J(x)$ are linearly dependent, or when $J(x)$ has high condition number \cite{matrixcomputations2013}.
When $J(x)$ is rank-deficient the LLS problem does not have a unique solution. When $J(x)$ has high condition number, the solution becomes inaccurate. This leads to inaccurate NLS steps which may cause slow convergence. The constraint $\|Dp\|_2 < \Delta$ of trust-region algorithms attenuates these problems somewhat by preventing extremely large steps but convergence rate still suffers. Generally, convergence of iterative first-order NLS algorithms is slow when the local linear approximation does not match the underlying function well as a consequence of high curvature or ill-conditioning.

\section{Motivation}
So far, the problem of ill-conditioning in GP with local optimization has been ignored in GP and SR literature. However, since GP has no countermeasures to prevent badly parameterized expressions, we expect it to occur frequently. For instance, each expression with redundant parameters does not have a full rank Jacobian. A simple example could be $f(x,\theta) = \theta_1 x_1 \cdot \theta_2 x_2 + \theta_3$
with Jacobian: $J(x,\theta) = (\theta_2 x_1 x_2, \theta_1 x_1 x_2, 1)$. Here the first two columns of $J(x)$ are linearly dependent for any value of $\theta$. Expressions like this may occur frequently in GP.

\begin{figure}
   \centering
    \begin{tikzpicture}
  \def\ws{0.07}
  \def\hs{0.2}
  \node (5) at (\ws*21,\hs*-16.8) {+};
  \node (7) at (\ws*16,\hs*-21) {$\theta_0$}; 
  \node (8) at (\ws*26,\hs*-21) {*};
  \node (9) at (\ws*21,\hs*-25.2) {$\theta_1$};
  \node (10) at (\ws*31,\hs*-25.2) {/};
  \node (11) at (\ws*10,\hs*-29.4) {*};
  \node (12) at (\ws*52,\hs*-29.4) {+};
  \node (13) at (\ws*0,\hs*-33.6) {$\theta_2$};
  \node (14) at (\ws*10,\hs*-33.6) {Y};
  \node (15) at (\ws*20,\hs*-33.6) {*};
  \node (16) at (\ws*40,\hs*-33.6) {*};
  \node (17) at (\ws*65,\hs*-33.6) {*};
  \node (18) at (\ws*15,\hs*-37.8) {$\theta_3$};
  \node (19) at (\ws*25,\hs*-37.8) {Y};
  \node (20) at (\ws*35,\hs*-37.8) {/};
  \node (21) at (\ws*45,\hs*-37.8) {$\theta_7$};
  \node (22) at (\ws*55,\hs*-37.8) {$\theta_8$};
  \node (23) at (\ws*65,\hs*-37.8) {Y};
  \node (24) at (\ws*75,\hs*-37.8) {*};
  \node (25) at (\ws*30,\hs*-42) {$\theta_4$};
  \node (26) at (\ws*40,\hs*-42) {*};
  \node (27) at (\ws*70,\hs*-42) {$\theta_9$};
  \node (28) at (\ws*80,\hs*-42) {Y};
  \node (29) at (\ws*30,\hs*-46.2) {$\theta_5$};
  \node (30) at (\ws*40,\hs*-46.2) {X};
  \node (31) at (\ws*50,\hs*-46.2) {*};
  \node (32) at (\ws*45,\hs*-50.4) {$\theta_6$};
  \node (33) at (\ws*55,\hs*-50.4) {X};
  \draw (5) -- (7);
  \draw (5) -- (8);
  \draw (8) -- (9);
  \draw (8) -- (10);
  \draw (10) -- (11);
  \draw (10) -- (12);
  \draw (11) -- (13);
  \draw (11) -- (14);
  \draw (11) -- (15);
  \draw (12) -- (16);
  \draw (12) -- (17);
  \draw (15) -- (18);
  \draw (15) -- (19);
  \draw (16) -- (20);
  \draw (16) -- (21);
  \draw (17) -- (22);
  \draw (17) -- (23);
  \draw (17) -- (24);
  \draw (20) -- (25);
  \draw (20) -- (26);
  \draw (24) -- (27);
  \draw (24) -- (28);
  \draw (26) -- (29);
  \draw (26) -- (30);
  \draw (26) -- (31);
  \draw (31) -- (32);
  \draw (31) -- (33);
    \end{tikzpicture}
    \caption{An overparameterized expression tree produced by genetic programming.}
    \label{fig:expr-tree}
 \end{figure}
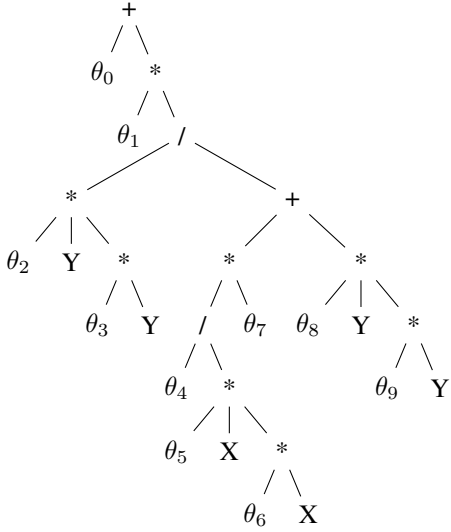

A concrete example is the expression tree shown in Figure~\ref{fig:expr-tree}. This expression was generated with tree-based genetic programming for the Pagie problem described in more detail below. The maximum size limit used in this run was relatively small with only 15 nodes\footnote{Operon counts a variable with a multiplicative coefficient as one node. Each node with a variable always has a coefficient.}. 
%
%
 The corresponding expression is
 \begin{equation}
   \theta_0 + \theta_1 \frac{\theta_2 Y \theta_3 Y} {\frac{\theta_4} {\theta_5 X \theta_6 X} \theta_7 + \theta_8 Y \theta_9 Y},
   \label{eqn:original}
 \end{equation}
which is obviously overparameterized with ten parameters that were optimized with NLS.
Automatic algebraic simplification of the expression leads to the simplified form
 \begin{equation}
   \theta_0' - \frac{\theta_1' Y^2}{ \theta_2' Y^2 - \frac{\theta_3'}{X^2}}
   \label{eqn:simplified}
 \end{equation}
 with only four parameters. However, this form is still overparameterized and should be re-parameterized to lead to:
  \begin{equation}
    \theta_0' - \frac{\theta_1' Y^2}{ Y^2 - \frac{\theta_2'}{X^2}} .
    \label{eqn:fixed}
 \end{equation}
It is important to note that the additional parameters in Equation ~\ref{eqn:original} do not allow a better fit than the form with only three parameters. The effective degree of freedom of both expressions is the same. Therefore the redundant parameters do not increase the danger of overfitting.
However, the over-parameterization has several other unwanted effects:
\begin{itemize}
  \item Reduced convergence speed of NLS
  \item Infinite number of solutions
  \item Parameters cannot be estimated accurately, leading to wide confidence intervals of parameters.
\end{itemize}

A comparison of the performance of NLS for Equation~\ref{eqn:original} and Equation~\ref{eqn:fixed} clearly demonstrates the problem. For Equation~\ref{eqn:original} the average number of function evaluations is $38$ and the average number of Jacobian evaluations is $31$ in NLS. The results are much better for the simplified Equation~\ref{eqn:fixed} with $12.5$ function and $8.5$ Jacobian evaluations on average. There was no difference in the success rate however. NLS converged to the best local optimum in $49\%$ (original equation) and $48\%$ (simplified equation) of the runs. In this experiment we used random 1000 restarts where the parameter vector was perturbed randomly around the local optimum. 

To get a better understanding of the prevalence of the issue we analyse the conditioning of the Jacobian matrices that occur in the local optimization problems.


\section{Methods}
We use the open-source GP implementation Operon\footnote{https://github.com/heal-research/operon}\cite{Burlacu2020}, which uses Eigen\cite{eigenweb} and the Levenberg-Marquardt implementation MINPACK \cite{minpackuserguide} and extend the code to log additional results. We extend the function which provides the Jacobian to calculate the rank and the condition number of $J(x)$ in each iteration via a singular value decomposition (SVD) $J = U\Sigma V^T$.
The SVD provides $\Sigma = \text{diag}(\sigma_1, \ldots, \sigma_k)$ with singular values $\sigma_i  \leq \sigma_{i+1}, i=1\ldots k-1$ and we use $\Sigma$ to determine the numeric rank $r$ based on the number of singular values which are larger than $k\, \epsilon\, \sigma_1$ \cite{OLearyRust2013} where $\epsilon$ is the gap between 1.0 and the next larger floating point number\footnote{determined via \texttt{std::numeric\_limits<T>.epsilon()}}. 
Additionally, we determine the condition number of $\kappa(J(x)) = \kappa(\Sigma) = \sigma_1 / \sigma_k$ as well as the condition number of the truncated SVD $\kappa(\Sigma_r) = \sigma_1 / \sigma_r$.
We calculate these values for each Jacobian generated for a solution candidate and use the minimum $r$ and the maximum $\kappa(\Sigma)$ and $\kappa(\Sigma_r)$ because they represent the worst values detected for each solution candidate. Additionally, we log average values and percentiles over all solution candidates in each generation. 
The same values are also logged for the final solutions returned by Operon.

For the expression tree in Figure~\ref{fig:expr-tree} the numeric rank
determination finds that there are only three effective degrees of
freedom. This is detected solely based on the Jacobian without any
kind of symbolic simplification. For Equation~\ref{eqn:original} we get $\log_{10} \kappa(J(x)) = 17.4$ and $\log_{10} \kappa(\Sigma_r) = 1.328$.
The simplified form with four parameters (Eqn.~\ref{eqn:simplified}) is still ill-conditioned with $\log_{10} \kappa(J(x)) = 17.4$ and $\log_{10} \kappa(\Sigma_r) = 1.32$.
Equation~\ref{eqn:fixed} has  $\log_{10} \kappa(J(x)) = \log_{10} \kappa(\Sigma_r) = 1.8$.

We use a diverse set of six datasets including synthetic and real-world datasets with and without noise as well as different dataset sizes and function complexities. Table \ref{tab:datasets} shows the characteristics of the datasets. These datasets are taken from the Operon github repository except for the Tower dataset which is described in \cite{vladislavleva2008order}.

\begin{table}
  \centering
  \begin{tabular}{lcccccc}
    Name & Type & d & n & Noise & Func. & Ref. \\
    \hline
    Airfoil & real & 5 & 100 & yes & - & \cite{brooks1989airfoil} \\
    Kotanchek & synthetic & 2 & 100 & no & \eqref{eq:vlad-1} & \cite{vladislavleva2008order} \\ 
    Pagie & synthetic & 2 & 676 & no & \eqref{eq:pagie} & \cite{pagie1997} \\
    Poly-10 & synthetic & 10 & 250 & no & \eqref{eq:poly10} & \cite{poli2006running} \\
    Salustowicz2D & synthetic & 2 & 600 & no & \eqref{eq:vlad-3} & \cite{vladislavleva2008order} \\ 
    Tower & real & 23 & 3136 & yes & - & \cite{vladislavleva2008order} \\
  \end{tabular}
  \caption{List of datasets}
  \label{tab:datasets}
\end{table}

\begin{align}
  f_1(x_1,x_2) & = \frac{\exp(- (x_1 - 1)^2)}{1.2 + (x_2 - 2.5)^2} \label{eq:vlad-1} \\
  f_2(x_1,x_2) & = \frac{1}{1 + x_1^{-4}} + \frac{1}{1 + x_2^{-4}} \label{eq:pagie} \\ 
  f_3(x) & = x_1 x_2 + x_3 x_4 + x_5 x_6 + x_1 x_7 x_9 + x_3 x_6 x_{10} \label{eq:poly10} \\
  f_4(x, y) & = e^{-x}  x^3 \cos x \sin x ( \cos x \sin(x)^2 - 1) (y - 5) \label{eq:vlad-3}
\end{align}

The GP parameters used in the experiments are shown in Table \ref{tab:parameters}. Our goal is to analyse the conditioning of local optimization problems and not to find the best model for each dataset. Therefore, we decided to set parameters to typical values close to the defaults of Operon. To study the effects of maximum size limits and the function set, we used four different maximum size values and a smaller and a larger function set\footnote{$aq(a,b) = a / \sqrt{1 + b^2}$ is the analytic quotient \cite{AQ}}. 
\begin{table}
  \begin{tabular}{ll}
    Parameter & Value \\
    \hline
    Population size & 1000 \\    
    Generations & 100 \\
    Local opt. iters. & 10\\
    Max. size & $\{15, 30, 50, 100\}$ nodes\\
    Initialization & BTC \cite{Burlacu2020} \\
    Recombination & Subtree crossover \\
    Mutation rate & 25\% \\
    Selection & Tournament group size 5 \\
    Replacement & Generational with one elite \\
    Function sets & Small $=\{+,\times, \div \}$ \\
    & Large $=\{+, \times, \div, \log |x|, \exp, \text{aq}(a, b), $\\
    & $\sin, \cos, \tanh, x^2, \sqrt{|x|}, \text{cbrt} \}$ \\
    Terminal set & variables with coefficients and numeric parameters
  \end{tabular}
  \caption{GP parameter settings}
  \label{tab:parameters}
\end{table}

For the analysis rank and condition number of solution candidates over generations we report the data gathered from a single run for each problem instance.

For the analysis of the rank and condition of the final solutions we execute 30 independent repetitions and analyse the average rank and condition number over the 30 solutions.

\section{Results}
In the following we summarize the key findings from the experiment and show plots for a few selected configurations. We do not show all plots because they are similar for all configurations. We use the nomenclature \emph{[Instance / Maximum size / Function set]} to refer to the different configurations.

\subsection{Conditioning of Solution Candidates}
\begin{figure}
  \centering
  \includegraphics[width=\columnwidth]{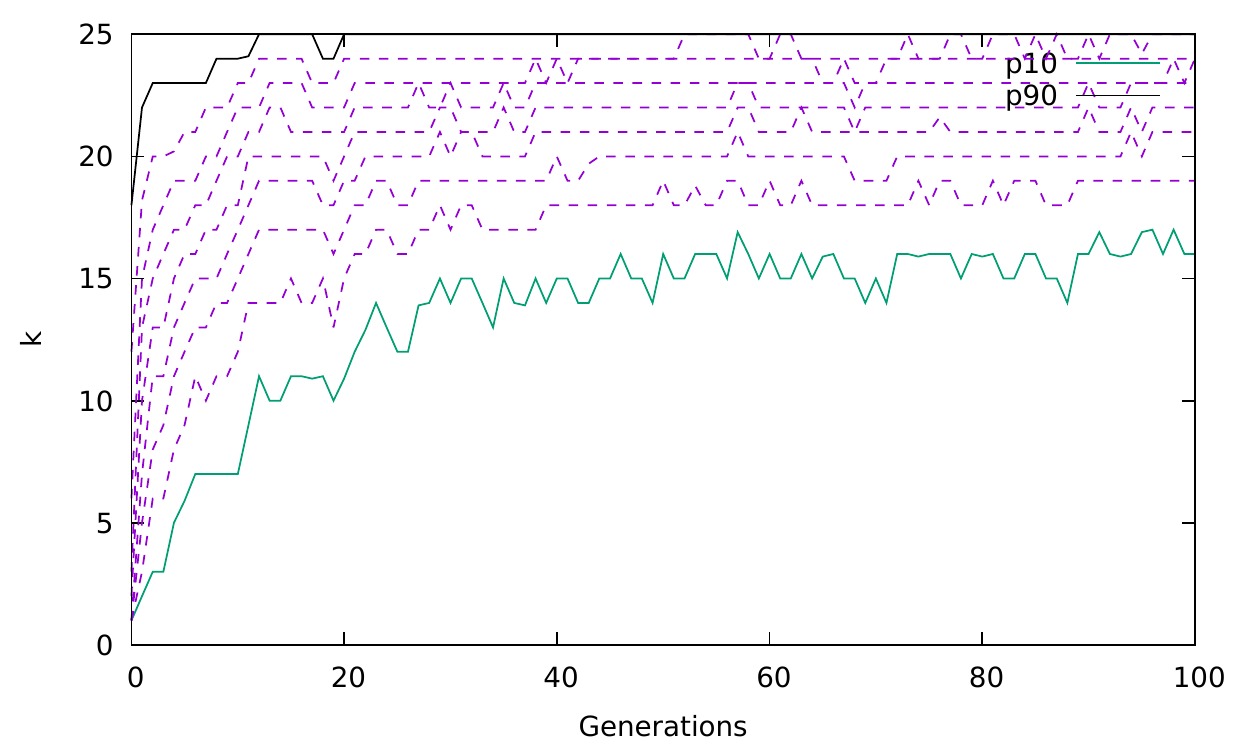}
  \includegraphics[width=\columnwidth]{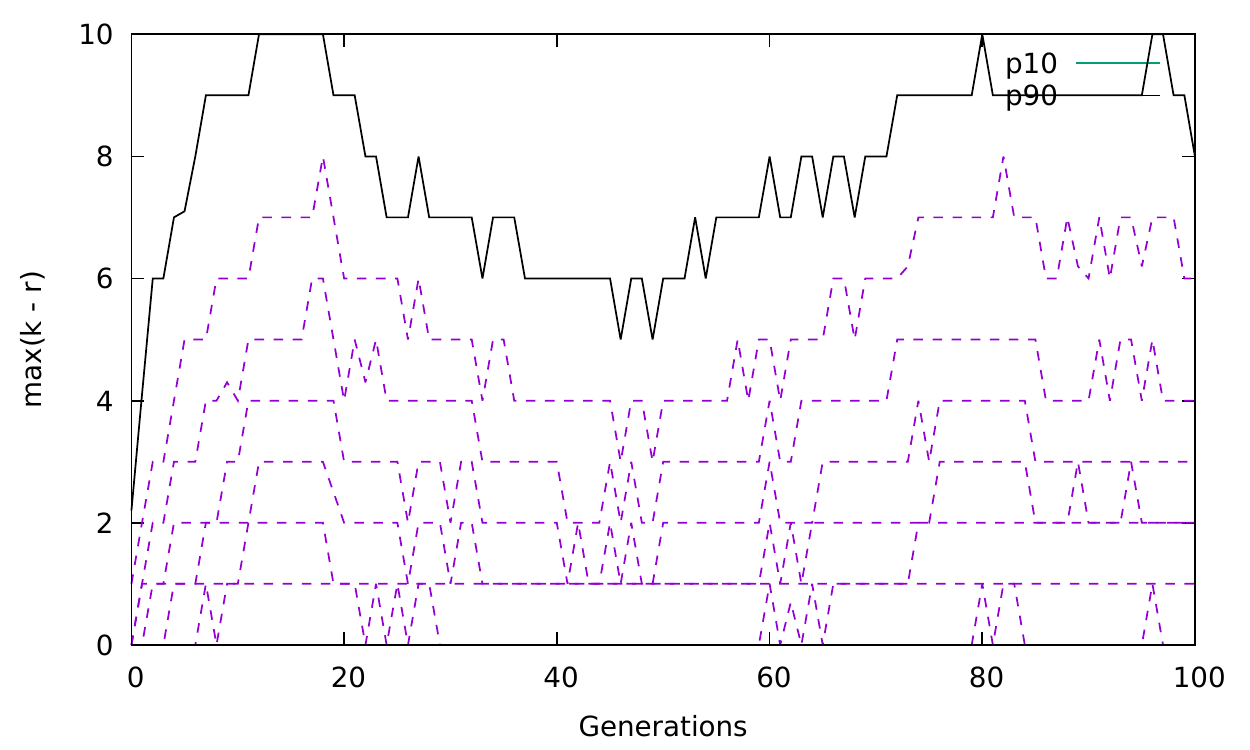}
  \caption{Distribution of the number of parameters $k$ and redundant parameters $k - r$ over generations for [Pagie-50-small]}
  \label{fig:pagie-50-small-param}
\end{figure}

Figure \ref{fig:pagie-50-small-param} shows for [Pagie-50-Small] the distribution of the number of parameters $k$ and the number of redundant parameters $k - r$ for all solution candidates over generations. The plots show that at the end of the run only 10\% of the solution candidates have less than 15 parameters and $k$ is  distributed between 15 to 25. 50\% of the solution candidates have two or more redundant parameters. This is in line with our hypothesis that many of the NLS problems in GP are rank-deficient.

\begin{figure}
  \centering
  \includegraphics[width=\columnwidth]{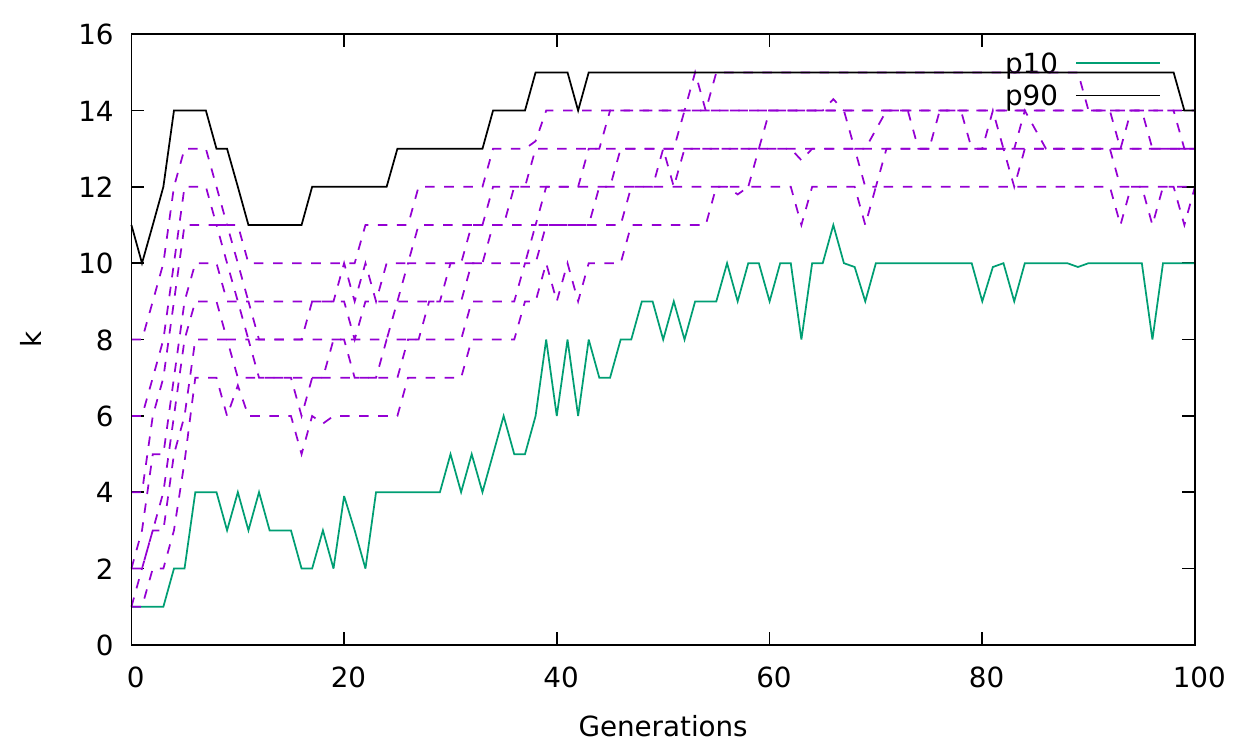}
  \includegraphics[width=\columnwidth]{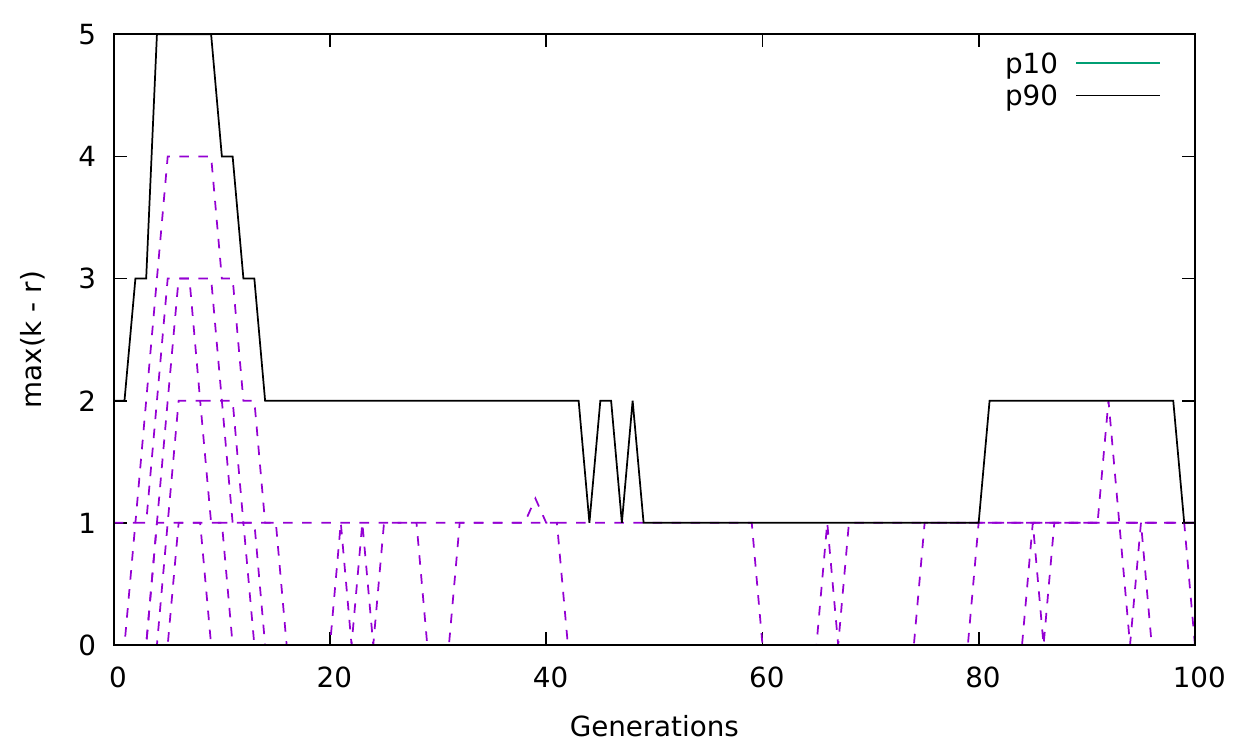}
  \caption{Distribution of the number of parameters $k$ and redundant parameters $k - r$ over generations for [Pagie-50-full]}
  \label{fig:pagie-50-full-param}
\end{figure}

As shown in Figure \ref{fig:pagie-50-full-param} the problem is less extreme when using the large function set because the larger number of univariate functions prohibits growth of trees and leads to fewer parameters. Additionally, the non-linear functions in the extended function set reduce the potential to produce linearly dependent parameters. For example the expression $\theta_0  \exp(\theta_1 x )$ does not have linearly dependent parameters. Another example are trigonometric functions $\theta_0 \cos (\theta_1 x + \theta_2)$. For logarithmic or inverse functions however, there may still be linear dependencies e.g., $\theta_0 \log (\theta_1 x + \theta_2) = \theta_0' \log(x + \theta_1'), \theta_1 > 0$. 

\begin{figure}
  \centering
  \includegraphics[width=\columnwidth]{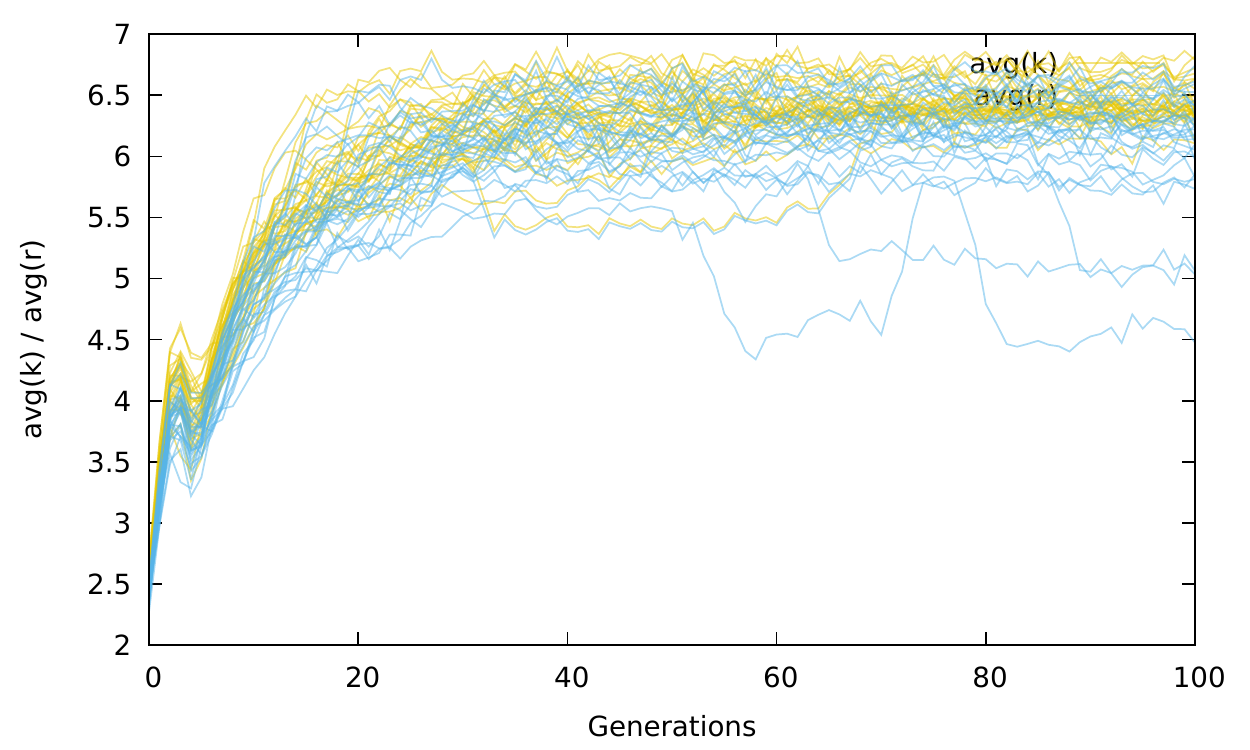}
  \includegraphics[width=\columnwidth]{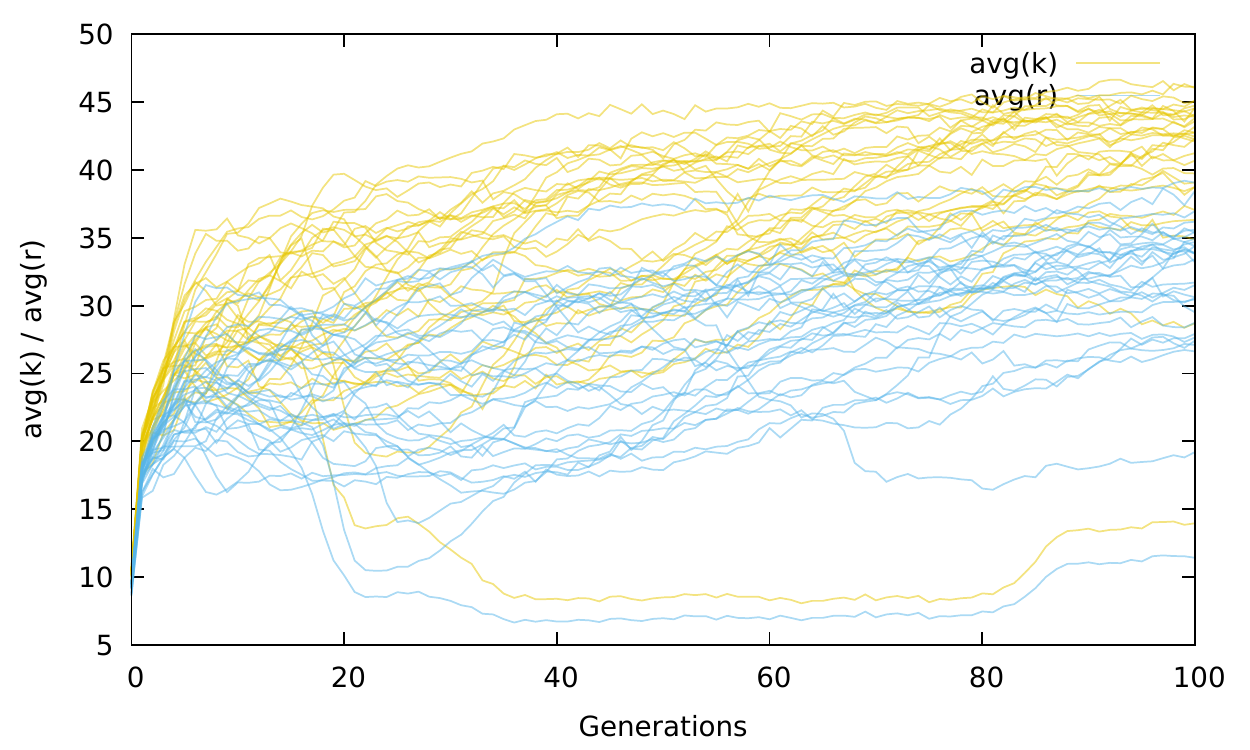}
  \caption{Average number of parameters and numeric rank over generations for 30 repetitions of [Pagie-15-small] (top) and [Pagie-100-small] (bottom)}
  \label{fig:pagie-15-100-small-rank}
\end{figure}

We observe a similar effect (smaller trees, fewer parameters, and fewer redundant parameters) for all six datasets when using the full function set instead of the small function set. Similarly, we observe the expected effect that increasing the maximum size limit increases the number of parameters. An important observation is that the number of redundant parameters also grows when the size limit is increased (see Figure \ref{fig:pagie-15-100-small-rank}). 

\begin{figure}
  \centering
  \includegraphics[width=\columnwidth]{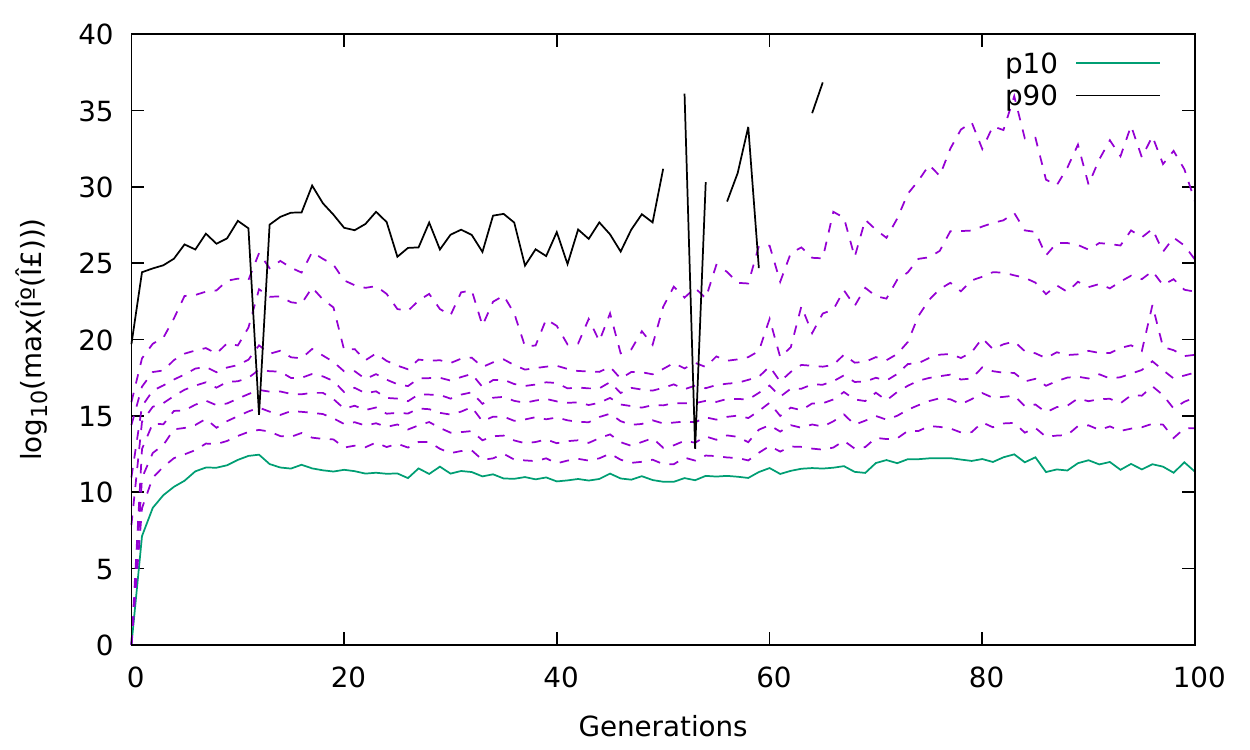}
  \includegraphics[width=\columnwidth]{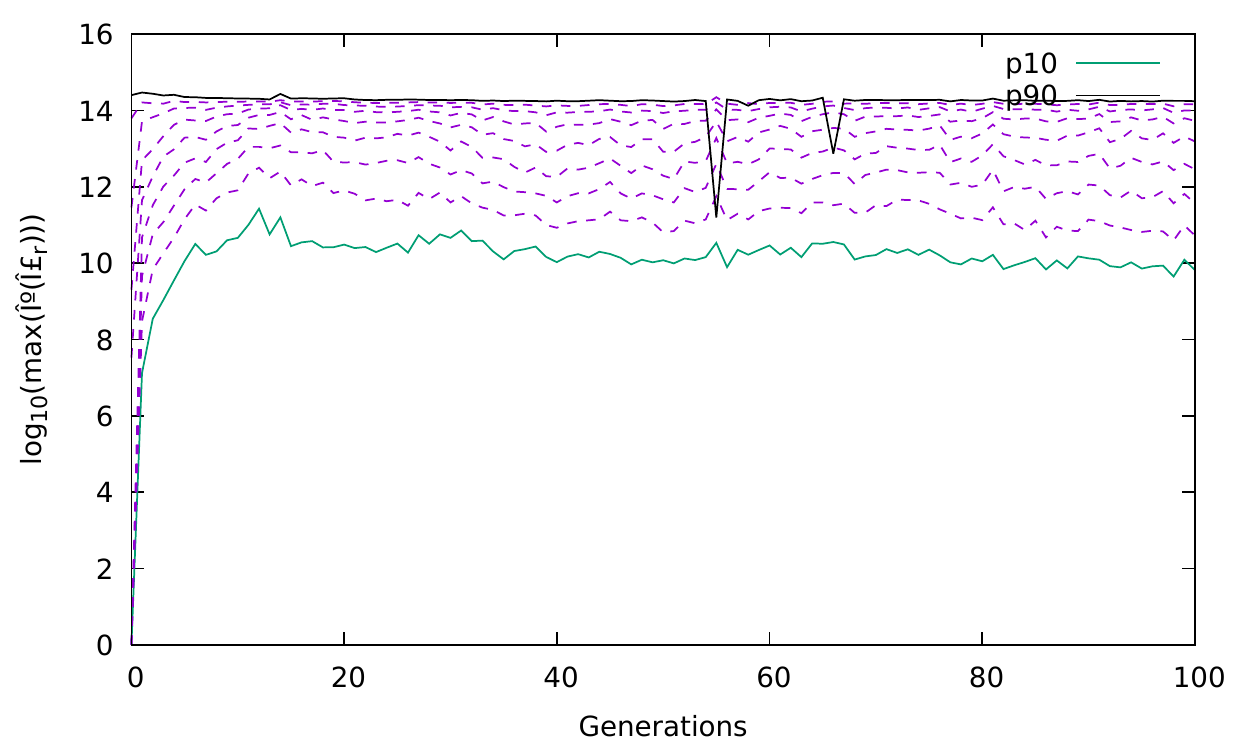}
  \caption{Distribution of the condition number $\max(\kappa(\Sigma))$ of the full Jacobian and of the truncated SVD $\max(\kappa(\Sigma_r))$ over generations for [Pagie-50-small]}
  \label{fig:pagie-50-small-cond}
\end{figure}

Figure \ref{fig:pagie-50-small-cond} shows the condition numbers of the Jacobian for the solution candidates over generations and shows that most of the NLS problems are ill-conditioned. When we truncate the SVD to the numeric rank, the condition number is naturally bounded from above, but most of the problems remain ill-conditioned even after truncation. This highlights that very small singular values are frequent in the Jacobian, which indicates that many expressions visited by GP have parameters with almost no effect. Only for 10\% of the solution candidates $\kappa(\Sigma)$ and $\kappa(\Sigma_r)$ are less than $10^{10}$. 

\begin{figure}
  \centering
  \includegraphics[width=\columnwidth]{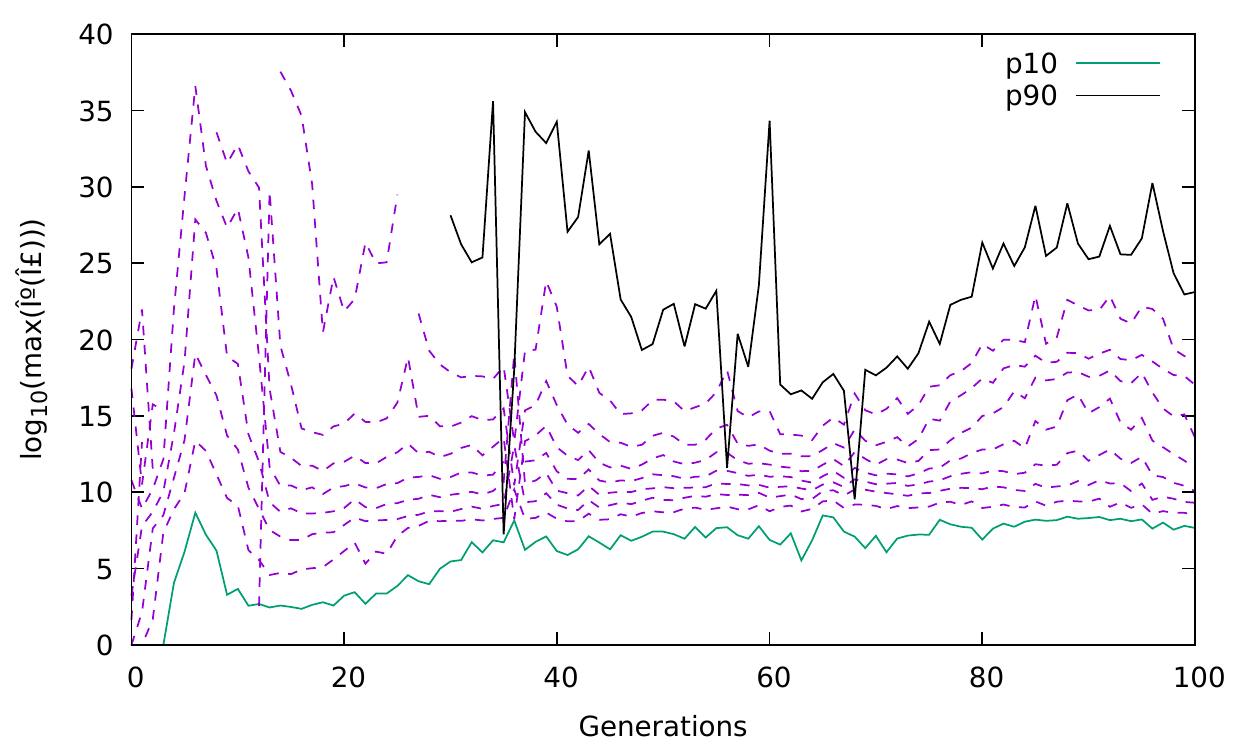} \includegraphics[width=\columnwidth]{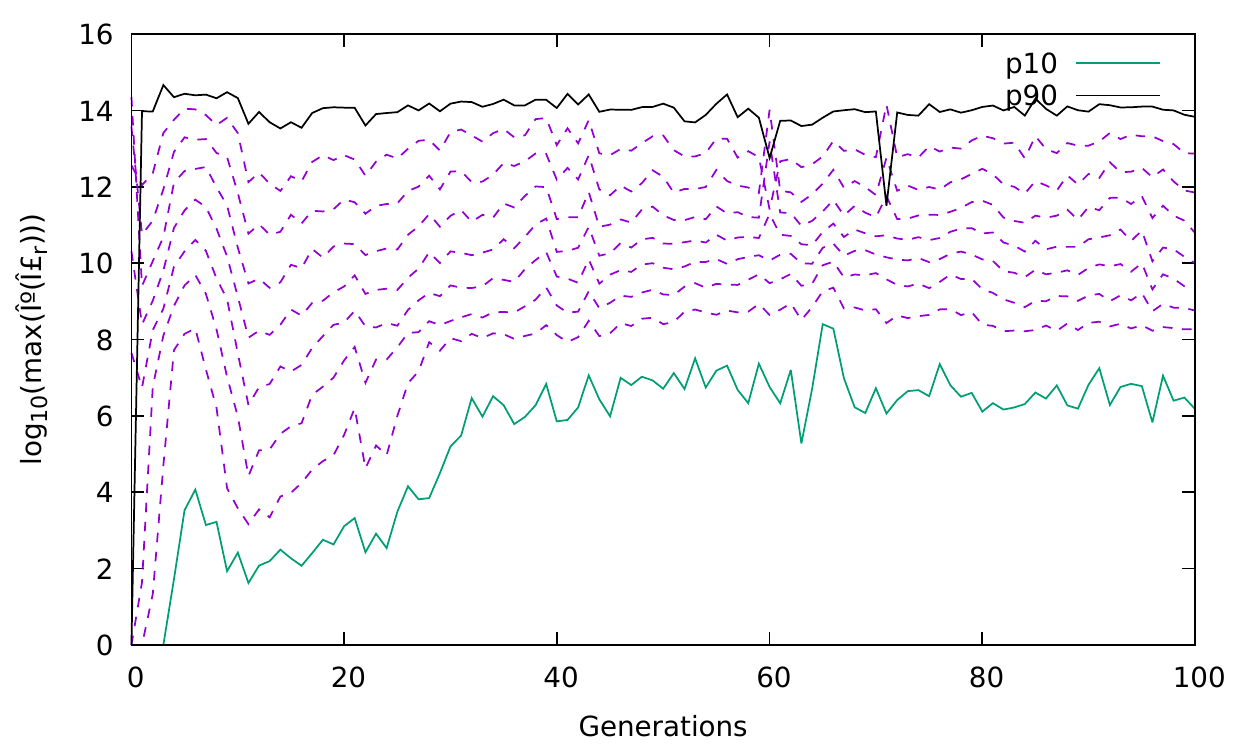}
  \caption{Distribution of the condition number $\max(\kappa(\Sigma))$ of the full Jacobian and of the truncated SVD $\max(\kappa(\Sigma_r))$ over generations for [Pagie-50-full]}
  \label{fig:pagie-50-full-cond_subset}
\end{figure}

For the full function set the problem is less extreme but we still observe a large fraction of high condition numbers in Figure~\ref{fig:pagie-50-full-cond_subset}.

So far, we have only shown plots from a single run. Figure~\ref{fig:pagie-15-100-small-rank} shows avg($k$) and avg($r$) of
solution candidates over generations for 30 runs. This shows that
avg($k$) is very similar over all 30 runs. The plots show that when
limiting the size to 15 nodes, the average rank is close to the number
of parameters in all runs. However, when using 100 nodes as the limit,
the two values diverge. This might indicate that with smaller trees GP
must use the available parameters to fit the data, while with larger
limits GP the expressions become bloated, as only a fraction of all parameters is
required to get a good fit. This means that the numeric rank could
potentially be used as an indicator for unnecessarily large
expressions. We recommend analysing this aspect in more detail in
future work.

\begin{table}
  \centering
  \begin{tabular}{lrrrrr}
    Dataset & max. size & $k$ & $k - r$ & $\kappa$ & $\kappa_r$ \\
    \hline
    Airfoil & 15        & 8    & 0     & 12  & 11 \\
    Airfoil & 30        & 15   & 0     & 14  & 13 \\
    Airfoil & 50        & 25   & 1     & 15  & 13 \\
    Airfoil & 100       & 49   & 7.5   & 22  & 14 \\
    Kotanchek & 15      & 8    & 0     & 10  & 10 \\
    Kotanchek & 30      & 15   & 0     & 12  & 12 \\
    Kotanchek & 50      & 25   & 0     & 13  & 12 \\
    Kotanchek & 100     & 47.5 & 3     & 18  & 13 \\
    Pagie & 15          & 8    & 0     & 10  & 10 \\
    Pagie & 30          & 15   & 0     & 12  & 12 \\
    Pagie & 50          & 25   & 1.5   & 17  & 12 \\
    Pagie & 100         & 46   & 3     & 19  & 13 \\
    Poly-10 & 15        & 8    & 0     & 9   & 9  \\
    Poly-10 & 30        & 15   & 0     & 12  & 12 \\
    Poly-10 & 50        & 25   & 0     & 14  & 13 \\
    Poly-10 & 100       & 49   & 5     & 18  & 14 \\
    Salustowicz2D & 15  & 8    & 0     & 9   & 9  \\
    Salustowicz2D & 30  & 15   & 0     & 12  & 12 \\
    Salustowicz2D & 50  & 25   & 0     & 12  & 12 \\
    Salustowicz2D & 100 & 48   & 3.5   & 19  & 13 \\
    Tower & 15          & 8    & 0     & 12  & 12 \\
    Tower & 30          & 15   & 0     & 12  & 11 \\
    Tower & 50          & 25   & 0     & 12  & 12 \\
    Tower & 100         & 49   & 3     & 19  & 13 \\      
  \end{tabular}
  \caption{Median number of parameters $k$, redundant parameters $k-r$, condition number $\kappa$, and truncated condition number $\kappa_r$ of 30 GP solutions with the small function set.}
  \label{tab:results-small}
\end{table}
\begin{table}
  \centering
  \begin{tabular}{lrrrrr}
    Dataset & max. size & $k$ & $k - r$ & $\kappa$ & $\kappa_r$ \\
    \hline
Airfoil & 15        & 6    & 0       & 14  & 12 \\
Airfoil & 30        & 12   & 2       & 17  & 13 \\
Airfoil & 50        & 18   & 2       & 23  & 13 \\
Airfoil & 100       & 20   & 1.5     & $\infty$    & 13 \\
Kotanchek & 15      & 6    & 0       & 2   & 2  \\
Kotanchek & 30      & 12   & 0       & 7   & 7  \\
Kotanchek & 50      & 19   & 0       & 10  & 9  \\
Kotanchek & 100     & 23.5 & 0       & 12  & 11 \\
Pagie & 15          & 4    & 0       & 5   & 4  \\
Pagie & 30          & 9    & 0       & 4   & 4  \\
Pagie & 50          & 15   & 0       & 12  & 10 \\
Pagie & 100         & 20.5 & 1       & 15  & 11 \\
Poly-10 & 15        & 7    & 0       & 9   & 9  \\
Poly-10 & 30        & 14   & 0       & 11  & 11 \\
Poly-10 & 50        & 22   & 0       & 12  & 12 \\
Poly-10 & 100       & 40   & 1       & 15  & 13 \\
Salustowicz2D & 15  & 7    & 0       & 3   & 3  \\
Salustowicz2D & 30  & 13   & 0       & 4   & 4  \\
Salustowicz2D & 50  & 19.5 & 0       & 10  & 10 \\
Salustowicz2D & 100 & 31.5 & 0       & 13  & 12 \\
Tower & 15          & 7    & 1       & 15  & 12 \\
Tower & 30          & 13   & 1       & 15  & 12 \\
Tower & 50          & 21   & 1       & 19  & 13 \\
Tower & 100         & 24.5 & 2       & 25  & 13 \\
  \end{tabular}
  \caption{Median number of parameters $k$, redundant parameters $k-r$, condition number $\kappa$, and truncated condition number $\kappa_r$ of 30 GP solutions with the large function set.}
  \label{tab:results-large}
\end{table}

\subsection{Conditioning of Final Solutions}
Tables \ref{tab:results-small} and \ref{tab:results-large} show median values for the final solutions produced in the 30 repetitions\footnote{Operon adds linear scaling coefficients before returning the final solution. The results are for the final expression before linear scaling.}. The number of parameters grows with size limits and is smaller for the large function set. A large fraction of final solutions has redundant (linearly dependent) parameters, as apparent in the column $k-r$. For smaller size limits and the large function set the median number of redundant parameters is smaller. The median condition numbers are very high even after truncation to the numeric rank. For the large function set, the condition numbers are smaller especially for smaller size limits.

\section{Summary and Conclusions}
Through the singular value decomposition, we calculated the numeric
rank and the condition number of the Jacobian matrices of nonlinear
least squares optimization problems occurring in genetic programming
with local optimization for symbolic regression.  The results
corroborate our hypothesis that rank-deficient and ill-conditioned
Jacobian matrices occur frequently in this scenario. Mainly, we
observed very high condition numbers for all problem instances even
after truncating the SVD to the numeric rank. Rank-deficiency and
ill-conditioning are more prevalent for the small function set ($\{+,
\times, \div\}$) and for large size limits.

These issues can severely limit the effectiveness and convergence rate
of nonlinear least squares algorithms used in many GP systems. In
light of our results we recommend considering adaptations to the
memetic local optimization techniques in GP. For instance, it would be
possible to optimize only the (projected) parameters with the largest
singular values or a random subset of parameters with several
restarts. The main aim should be to reduce the condition number of the
Jacobian.

We used only a single GP implementation for our experiments but we
expect that the problem affects other GP implementations as well
because the main cause for ill-conditioning and rank-deficiency is
that GP assembles expressions without considering these aspects at
all. However, the issue may be exaggerated for Operon because it enforces multiplicative coefficients for all variable nodes.

Symbolic regression algorithms that use a restricted encoding or
a limited grammar that limits overparameterization or bad scaling are
likely to be less affected (e.g. \cite{ITEA}). Analysis for these algorithms would be interesting as a follow-up paper.

In any case we recommend to check the conditioning of the produced solutions because it is easy to accomplish via the SVD.

Coincidently, we observed that numeric rank determination allows to
detect overparameterized expressions and could therefore be useful to
detect bloated solution candidates. Numeric rank determination could
also be useful for simplification of GP solutions.  These ideas should also be followed up in future research.

\section*{Acknowledgment}
G.K. acknowledges support by the Christian Doppler Research
Association and the Austrian Federal Ministry of Digital and Economic
Affairs within the Josef Ressel Center for Symbolic Regression.
The author thanks the maintainers of the GP bibliography for providing such a well-curated BibTeX database.

\bibliographystyle{IEEEtran}
\bibliography{references}

\end{document}